\documentclass[runningheads]{llncs}
\usepackage{graphicx}

\usepackage{tikz}
\usepackage{comment}
\usepackage{amsmath,amssymb} 
\usepackage{color}

\usepackage[accsupp]{axessibility}  

\usepackage[width=122mm,left=12mm,paperwidth=146mm,height=193mm,top=12mm,paperheight=217mm]{geometry}

\usepackage[pagebackref=true,breaklinks=true,letterpaper=true,colorlinks,bookmarks=false]{hyperref}
\usepackage{multirow}
\usepackage{subfigure}
\usepackage{float}
\graphicspath{{pics/}}
\usepackage{url}
\usepackage{algorithm}
\usepackage{algorithmic}

\begin{document}
\pagestyle{headings}
\mainmatter
\def\ECCVSubNumber{1751}  

\title{INT: Towards Infinite-frames 3D Detection with An Efficient Framework} 

\titlerunning{INT: Towards Infinite-frames 3D Detection}
%
\author{Jianyun Xu \and Zhenwei Miao$^{\dag}$ \and Da Zhang \and Hongyu Pan \and Kaixuan Liu
\and Peihan Hao \and Jun Zhu \and Zhengyang Sun \and Hongmin Li \and Xin Zhan}
\authorrunning{Jianyun Xu et al.}
%
\institute{Alibaba Group \\
\email{\{xujianyun.xjy,zhenwei.mzw\}@alibaba-inc.com $^{\dag}$corresponding author}}
\maketitle

\begin{abstract}

It is natural to construct a multi-frame instead of a single-frame 3D detector for a continuous-time stream. Although increasing the number of frames might improve performance, previous multi-frame studies only used very limited frames to build their systems due to the dramatically increased computational and memory cost. To address these issues, we propose a novel on-stream training and prediction framework that, in theory, can employ an infinite number of frames while keeping the same amount of computation as a single-frame detector. This {\bf in}fini{\bf t}e framework (INT), which can be used with most existing detectors, is utilized, for example, on the popular CenterPoint, with significant latency reductions and performance improvements. We've also conducted extensive experiments on two large-scale datasets, nuScenes and Waymo Open Dataset, to demonstrate the scheme's effectiveness and efficiency. By employing INT on CenterPoint, we can get around 7\% (Waymo) and 15\% (nuScenes) performance boost with only 2\~{}4ms latency overhead, and currently SOTA on the Waymo 3D Detection leaderboard. Code is available at: \href{https://github.com/ADLab-AutoDrive/INT}{https://github.com/ADLab-AutoDrive/INT}.
\end{abstract}

\section{Introduction}

3D object detection from pointclouds has been proven as a viable robotics vision solution, particularly in autonomous driving applications. Many single-frame 3D detectors~\cite{2019PointRCNN,2019votenet,2018VoxelNet,2019PointPillars,2018SECOND,2021Center,2021AFDetV2,2020PV} are developed to meet the real-time requirement of the online system. Nevertheless, it is more natural for a continuous-time system to adopt multi-frame detectors that can fully take advantage of the time-sequence information. However, as far as we know, few multi-frame 3D detectors are available for long frame sequences due to the heavy computation and memory burden. It is desirable to propose a concise real-time long-sequence 3D detection framework with promising performance.

\begin{figure*}[t]
    \centering
    \includegraphics[width=0.95\linewidth]{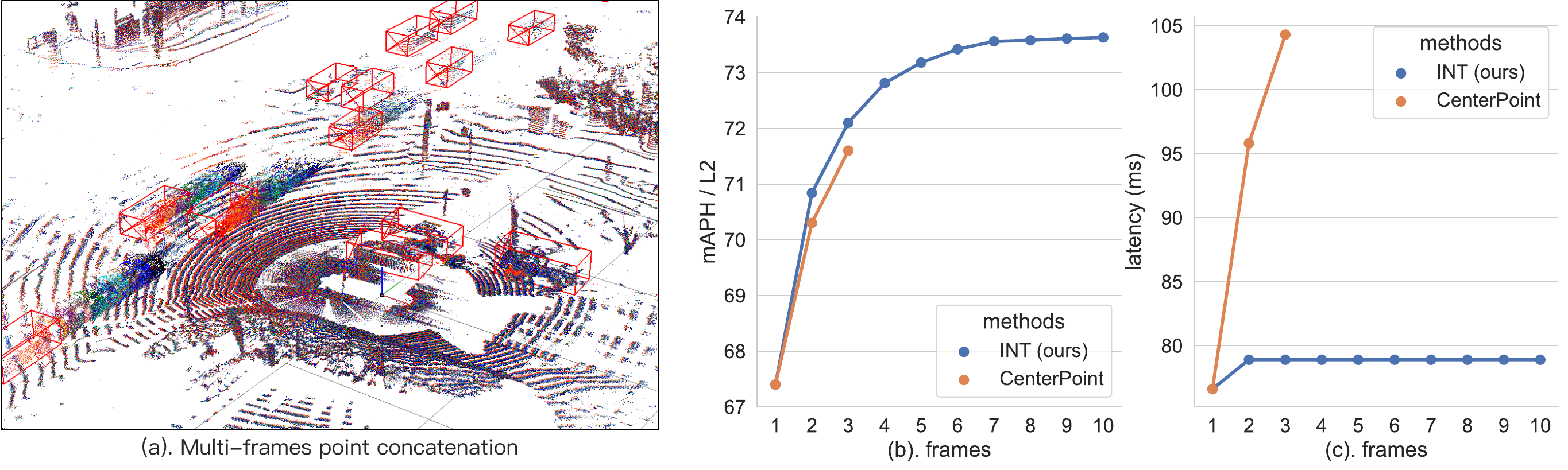}
    \caption{Impact of frames used in detectors on Waymo {\it val} set. While CenterPoint's performance improves as the number of frames grows, the latency also increases dramatically. On the other hand, our INT keeps the same latency while increasing frames.}
    \label{fig:frames-exp-waymo}
\end{figure*}

Existing works~\cite{2021Center,20214DNet,2020An,2020DVID,choy20194d,wu2020motionnet,yang20213D-MAN} demonstrate that multi-frame models yield performance gains over single-frame ones. However, these approaches require loading all the used frames at once during training, resulting in very limited frames being used due to computational, memory, or optimization difficulties.
Taking the SOTA detector CenterPoint~\cite{2021Center} as an example, it only uses two frames~\cite{2021Center} on the Waymo Open Dataset~\cite{sun2020scalability}. While increasing the number of frames can boost the performance, it also leads to significant latency burst as shown in Fig.~\ref{fig:frames-exp-waymo}.
Memory overflow occurs if we keep increasing the frames of CenterPoint, making both training and inference impossible.
As a result, we believe that the number of frames used is the bottleneck preventing multi-frame development, and we intend to break through this barrier first.

There are two major problems that limit the number of frames in a multi-frame detector: 1) repeated computation. Most of the current multi-frame frameworks have a lot of repeated calculations or redundant data that causes computational spikes or memory overflow; 2) optimization difficulty. Some multi-frame systems have longer gradient conduction links as the number of frames increases, introducing optimization difficulties.

To alleviate the above problems, we propose INT (short for {\bf in}fini{\bf t}e), an on-stream system that theoretically allows training and prediction to utilize infinite number of frames. INT contains two primary components: 1) a Memory Bank (MB) for temporal information fusion and 2) a Dynamic Training Sequence Length (DTSL) strategy for on-stream training. The MB is a place to store the recursively updated historical information so that we don't have to compute past frames' features repeatedly. As a result, it only requires a small amount of memory but can fuse infinite data frames.
To tackle the problem of optimization difficulty, we truncate the gradient of back propagation to the MB during training. However, an inherent flaw of iteratively updating MB on a training stream is that historical and current information are not given by the same model parameters, leading to training issues. To solve this problem, DTSL is employed.
The primary idea of DTSL is to start with short sequences and quickly clear the MB to avoid excessive inconsistency between historical and current data; then gradually lengthen the sequence as training progresses since the gap between historical and current model parameters becomes negligible.


To make INT feasible, we propose three modules: SeqFusion, SeqSampler and SeqAug. SeqFusion is a module in MB for temporal fusion that proposes multiple fusion methods for two types of data commonly used in pointcloud detection, i.e., point-style and image-style. SeqSampler is a sequence index generator that DTSL uses to generate training sequences of different lengths at each epoch. Finally, SeqAug is a data augmentation for on-stream training, capable of maintaining the same random state on the same stream.

Our contributions can be summarized as:
\begin{itemize}
    \item We present INT, an on-stream multi-frame system made up of MB and DTSL that can theoretically be trained and predicted using infinite frames while consuming similar computation and memory as a single-frame system.
    \item We propose three modules, SeqFusion, SeqSampler and SeqAug, to make INT feasible.
    \item We conduct extensive experiments on nuScenes and Waymo Open Dataset to illustrate the effectiveness and efficiency of INT.
\end{itemize}


\section{Related Work}
\label{sec:related_work}
\subsection{3D Object Detection}
\label{subsec:related_work_single}

Recent studies on 3D object detection can be broadly divided into three categories: LiDAR-based~\cite{2019PointRCNN,2019votenet,2018VoxelNet,2019PointPillars,2018SECOND,2021Center,2021AFDetV2,li2016veloFCN,meyer2019lasernet,zhou2020mvf,xu2021rpvnet}, image-based~\cite{ku2019monocular,chong2022monodistill,wang2019pseudo,li2019gs3d}, and fusion-based~\cite{chen2017mv3d,liang2019MMF,qi2018frustum,vora2020pointpainting,yoo20203dcvf,20214DNet}. Here we focus on LiDAR-based schemes.

According to the views of pointclouds, 3D detectors can be classified as point-based, voxel-based, range-based, and hybrid. PointRCNN~\cite{2019PointRCNN} and VoteNet~\cite{2019votenet} are two representative point-based methods that use a structure like PointNet++~\cite{qi2017pointnet++} to extract point-by-point features. These schemes are characterized by better preservation of pointclouds' original geometric information. Still, they are sensitive to the number of pointclouds that pose serious latency and memory issues. In contrast, voxel-based solutions, such as VoxelNet~\cite{2018VoxelNet}, PointPillars~\cite{2019PointPillars}, Second~\cite{2018SECOND}, CenterPoint~\cite{2021Center} and AFDetV2~\cite{2021AFDetV2} are less sensitive to the number of pointclouds. They convert the pointcloud into 2D pillars or 3D voxels first, then extract features using 2D or 3D (sparse) convolution, making them easier to deploy. Another category is rangeview-based schemes~\cite{li2016veloFCN,meyer2019lasernet,2021RangeDet} that perform feature extraction and target prediction on an efficient unfolded spherical projection view. Meanwhile, they contend with target scale variation and occlusion issues, resulting in performance that is generally inferior to that of voxel-based schemes. Hybrid methods~\cite{2020PV,zhou2020mvf,xu2021rpvnet} attempt to integrate features from several views to collect complementing information and enhance performance.

We define two data styles according to the data format to facilitate the analysis of different detectors:
\begin{itemize}
  \item \textit{\textbf{Image-style}}. Well-organized data in 2D, 3D, or 4D dimensions similar to that of an image.
  \item \textit{\textbf{Point-style}}. Disorganized data such as pointclouds and sparse 3D voxels.
\end{itemize}

The data in pointcloud-based detectors, including input, intermediate feature, and output, is generally point-style or image-style. We design the fusion algorithms for both point-style and image-style data in INT's Memory Bank, so that INT can be employed in most 3D detectors. In this work, we choose the recently popular CenterPoint~\cite{2021Center} as the baseline for studies since it performs better in terms of efficiency and performance, and it is now scoring at SOTA level on the large-scale open datasets nuScenes~\cite{caesar2020nuscenes} and Waymo Open Dataset~\cite{sun2020scalability}.

\subsection{Multi-frame Methods}
\label{subsec:related_work_multi}
There have been a variety of LSTM-based techniques in the field of video object detection, such as~\cite{feng2019vod1,kang2017vod2,xiao2018vod3}. Transformer-based video detection schemes have recently emerged~\cite{he2021transvod}, however transformer may not be suitable for working on-stream because it naturally needs to compute the relationship between all frames, which implies a lot of repeated computations.

\begin{figure*}[t]
    \centering
    \includegraphics[width=0.98\linewidth]{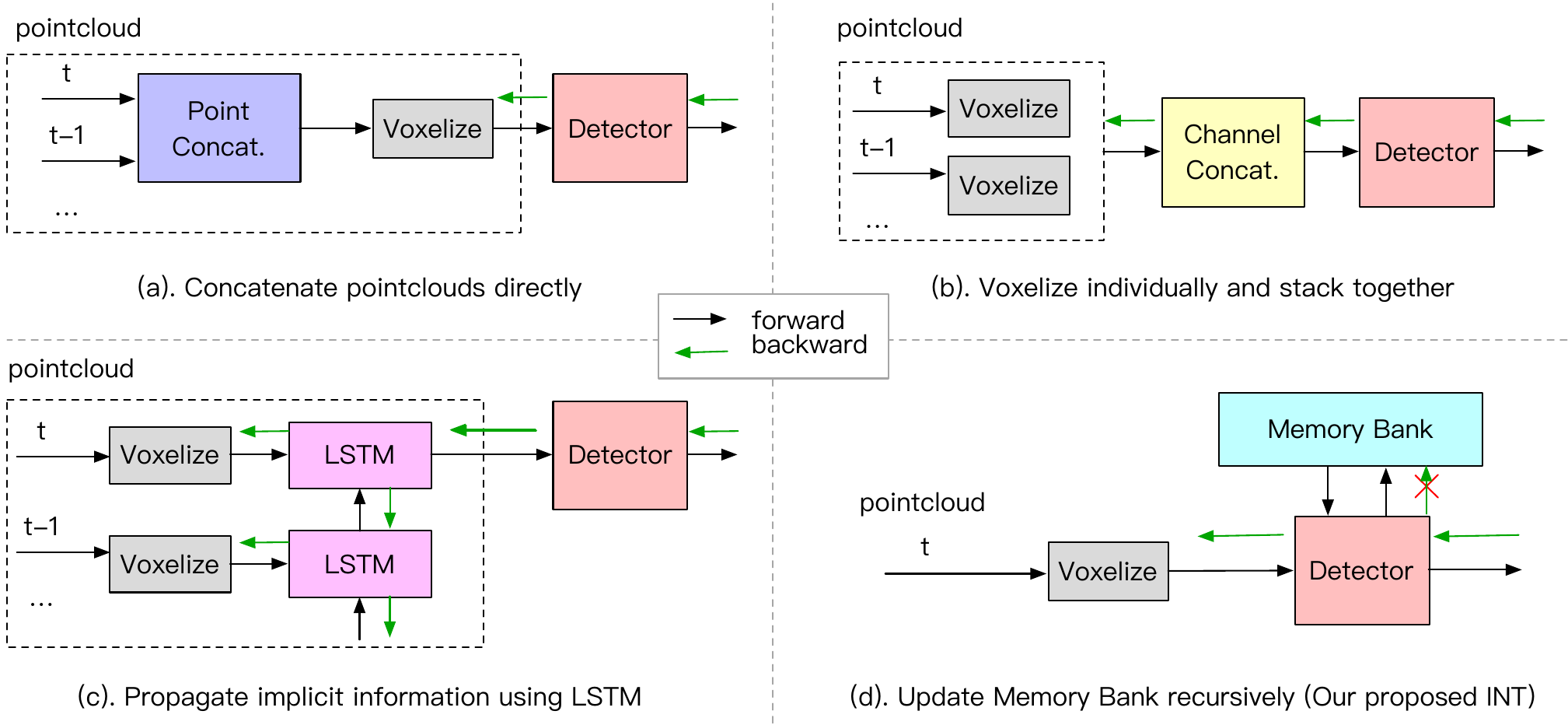}
    \caption{Training phase of different multi-frame schemes. Operations inside dash rectangle either involve repetitive computation or raise memory burden, which leads to a very limited frame number for training.}
    \label{fig:MF-Compare}
\end{figure*}

Recent methods for multi-frame pointclouds can be roughly divided into three categories as shown in Fig.~\ref{fig:MF-Compare}(a), (b) and (c). \cite{caesar2020nuscenes,20214DNet,2018SECOND} directly concatenate multi-frame pointclouds and introduce a channel indicating their relative timestamps, as shown in Fig.~\ref{fig:MF-Compare}(a). While this method is simple and effective, it involves a lot of unnecessary computations and increases the memory burden, making it unsuitable for more frames. Instead of merging at the point level, MinkowskiNet~\cite{choy20194d} and MotionNet~\cite{wu2020motionnet} combine multiple frames at the feature map level. They must voxelize multi-frame pointclouds independently before stacking the feature maps together and extracting spatio-temporal information using 3D or 4D convolution, as depicted in Fig.~\ref{fig:MF-Compare}(b). Obviously, this approach requires repeated data processing and is memory intensive, thus the number of frames is very limited.



To overcome above difficulties, 2020An~\cite{2020An} and 3DVID~\cite{2020DVID} proposed LSTM or GRU-based solutions and solve the computational and memory issues in the inference phase. However, the gradient transfer to the history frames still results in a considerable memory overhead and introduce optimization difficulties during training, so the number of frames cannot be high, as shown in Fig.~\ref{fig:MF-Compare}(c). To handle the problem more thoroughly, we propose computing the gradient for the current data and not for the historical data during training, as shown in Fig.~\ref{fig:MF-Compare}(d). We then employ a Dynamic Training Sequence Length (DTSL) strategy to eliminate the potential information inconsistency problem. Similarly, 3D-MAN~\cite{yang20213D-MAN} stores historical information in a Memory Bank that does not participate in the gradient calculation. However, 3D-MAN needs to store a fixed number of frames of historical proposals and feature maps, which increases the amount of memory required for its training as the number of frames increases. To get around this problem, we propose recursively updating the Memory Bank's historical information.

To the best of our knowledge, we are the first 3D multi-frame system that can be trained and inferred with infinite frames.

\section{Methodology}
\label{sec:Methodology}
We present INT framework in this section. The overall architecture is detailed in Sec.~\ref{sec:Framework}. Sec.~\ref{sec:SeqFusion} gives the sequence fusion (SeqFusion) methods of Memory Bank. Sec.~\ref{sec:SeqSampler} and~\ref{sec:SeqAug} illustrate the training strategies, including sequence sampler (SeqSampler) and sequence data augmentation (SeqAug), respectively.

\subsection{Overview of INT Framework}
\label{sec:Framework}
\begin{figure*}
    \centering
    \includegraphics[width=0.9\linewidth]{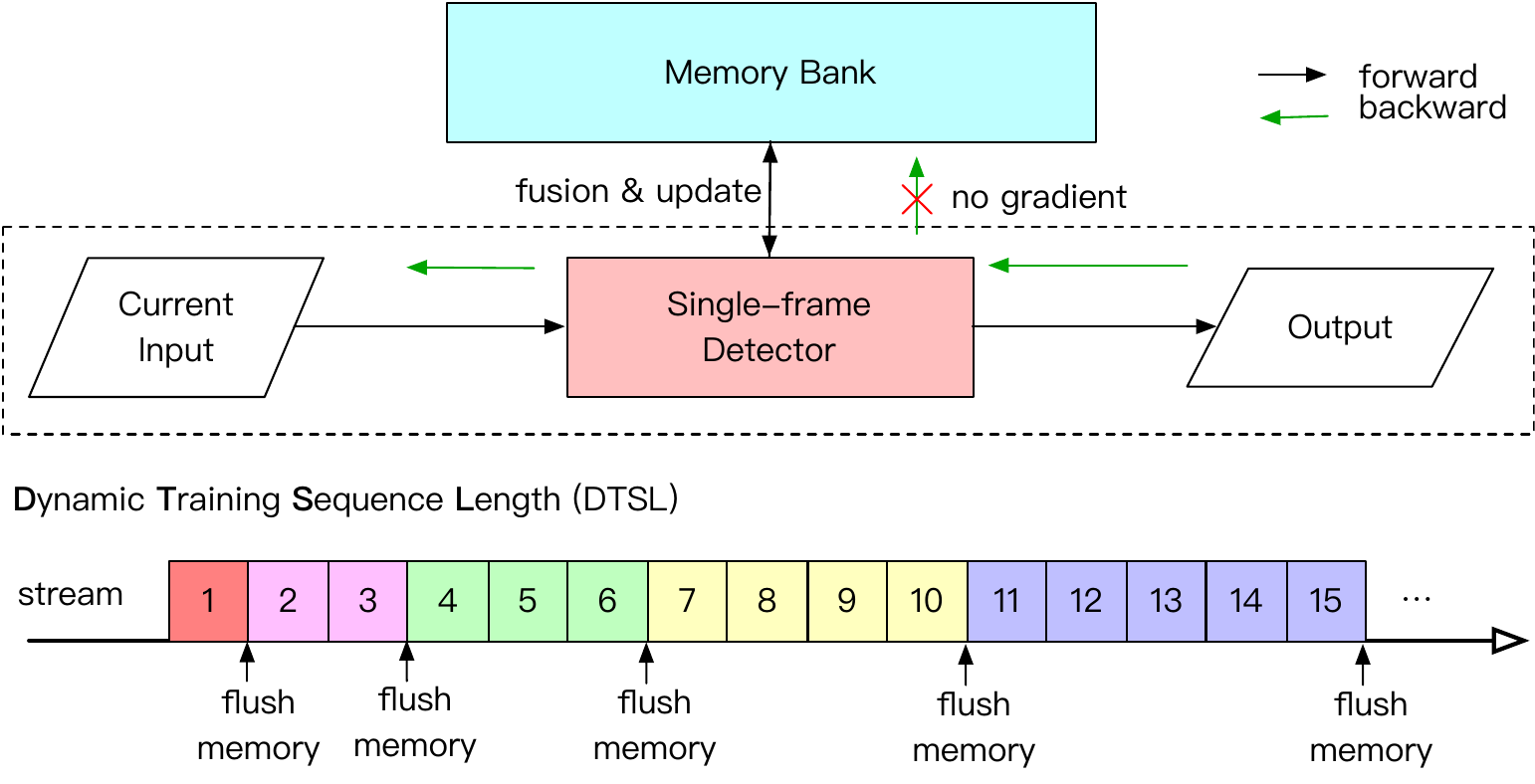}
    \caption{Overview of INT framework. It consists of a single-frame detector and a Memory Bank. The Dynamic Training Sequence Length below serves as a training strategy.}
    \label{fig:INT-Framework}
\end{figure*}

The INT framework in Fig.~\ref{fig:INT-Framework} is highly compact. The main body consists of a single-frame detector and a recursively updated Memory Bank (MB). The Dynamic Training Sequence Length (DTSL) below serves as a training strategy that is not needed for inference (inference only needs the pointclouds input in chronological order). In addition, there are no special requirements for single-frame detector selection. For example, any detector listed in Sec.~\ref{subsec:related_work_single} can be utilized.


\subsubsection{Memory Bank (MB).}
The primary distinction between INT and a regular multi-frame detector is the MB, which stores historical data so that we do not have to compute past features repeatedly. MB is comparable to the hidden state in LSTM while it is more flexible, interpretable, and customizable. The user has complete control over where and what information should be saved or retrieved. For example, in Sec.~\ref{sec:SeqFusion}, we show how to store and update several forms of data. Furthermore, we choose to update the MB recursively to solve the problem of excessive memory cost. To tackle the problem of optimization difficulty, we truncate the gradient of backpropagation to the MB during training.


\subsubsection{Dynamic Training Sequence Length (DTSL).}
\label{subsec:DTSL}
 A problem with INT training on stream is the information gained from the current observation is not derived using the same model parameters as the past data in the MB. This could lead to inconsistencies in training and prediction, which is one of the key reasons why prior multi-frame work was not trained on stream. To solve this problem, we offer the DTSL: beginning with a small sequence length and gradually increasing it, as indicated at the bottom of Fig.~\ref{fig:INT-Framework}. This is based on the following observation: as the number of training steps increases, model parameter updates get slower, and the difference in information acquired from different model parameters becomes essentially trivial. As a result, when the model parameters are updated quickly, the training sequence should be short so that the Memory Bank can be cleaned up in time. Once the training is stable, the sequence length can be increased with confidence.
DTSL could be defined in a variety of ways, one of which is as follows:
\begin{equation}\label{eq:dtsl}
    DTSL=max(1, \ \lfloor l_{max}\cdot min(1, \ max(0, \ 2\cdot \frac{ep_{cur}}{ep_{all}}-0.5)) \rfloor)
\end{equation}
where $l_{max}$ is the maximum training sequence length, $ep_{cur}$ and $ep_{all}$ is current epoch and total epoch number, respectively.

\subsection{SeqFusion}
\label{sec:SeqFusion}
Temporal fusion in the Memory Bank is critical in the INT framework. As the type of data in a 3D detector is either point-style or image-style, as indicated in Sec.~\ref{subsec:related_work_single}, we develop both the point-style and image-style fusion algorithms. In general, original pointcloud, sparse voxel, predicted object, etc., fall into the point-style category. Whereas dense voxel, intermediate feature map, final prediction map, etc., fall into the image-style category.

\subsubsection{Point-style fusion.}
\label{subsec:point_style}
Here we propose a general and straightforward practice: concatenating past point-style data with present data directly, using a channel to identify the temporal relationship. The historical point-style data is put into a fixed length FIFO queue, and as new observations arrive, foreground data is pushed into it, while oldest data is popped out. According to the poses of ego vehicle, the position information in the history data must be spatially transformed before fusion to avoid the influence of ego movement. The point-style fusion is formulated as:
\begin{align}
\centering
  &  T_{rel} = T_{cur}^{-1} \cdot T_{last},\\ \label{eq:Tr}
  &  P_f = PointConcat(P_{cur}, \ T_{rel}\cdot P_{last})
\end{align}
where $T_{last}$ and $T_{cur}$ are the last and current frame's ego vehicle poses, respectively, while $T_{rel}$ is the calculated relative pose between the two frames. $P_{last}$ refers to the past point-style data in Memory Bank and $P_f$ is the fused data of $P_{last}$ and current $P_{cur}$.

\begin{figure*}[]
    \centering
    \includegraphics[width=1.0\linewidth]{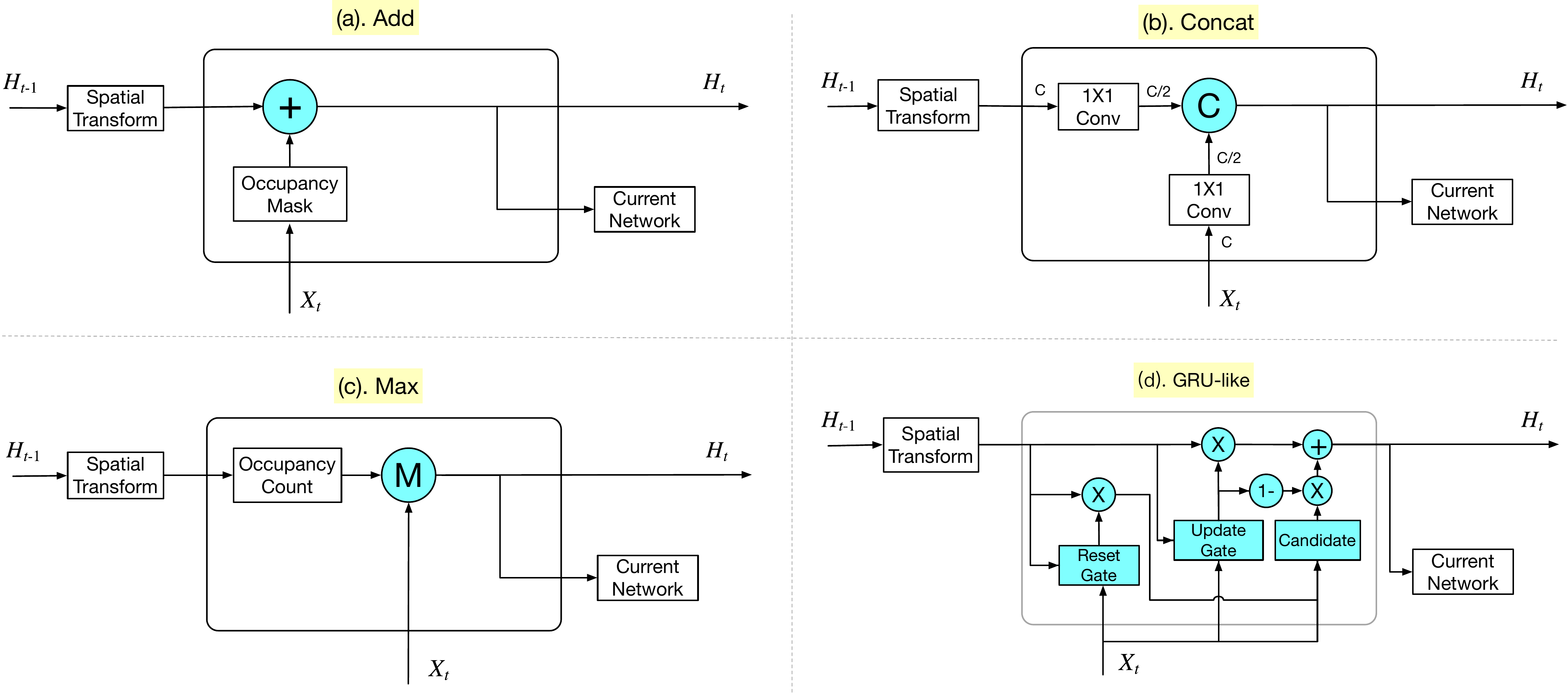}
    \caption{Four temporal fusion methods for image-style data. Occupancy Mask and Occupancy Count in {\it Add} and {\it Max} are used to distinguish different moments.}
    \label{fig:SeqFusion}
\end{figure*}

\subsubsection{Image-style fusion.}
\label{subsec:image_style}
We propose four fusion algorithms for the image-style data, including {\it Add}, {\it Max}, {\it Concat} and {\it GRU-like} as depicted in Fig.~\ref{fig:SeqFusion} (a), (b), (c) and (d). As the historical image-style data and current data should be identical in dimensions based on recursive updates, {\it Add} and {\it Max} are simple in design and implementation. The computational overhead of these two fusion methods is cheap. We also devise the {\it Concat} fusion approach, in which both the historical and the current feature channels are first compressed to 1/2 of the origin and then concatenated along channel dimension. To investigate the impact of long-term data, we develop a {\it GRU-like} fusion method with learnable parameters to select which data should be kept and which should be discarded. To eliminate the effect of ego vehicle motion, historical image-style data must be spatially transformed first. The image-style fusion process can be summarized as follows:
\begin{align}
\centering
  &  \tilde I_{last} = F_{sample}(I_{last}, \ F_{affine}(T_{rel}, \ s)),\\ \label{eq:image_Tr}
  &  I_f = Fusion(I_{cur}, \ \tilde I_{last})
\end{align}
where $T_{rel}$ is the same as Eq.~\ref{eq:Tr}. $F_{affine}(\cdot)$ and $F_{sample}(\cdot)$ refer to {\it affine grid} and {\it grid sample} operation respectively, which are proposed in~\cite{jaderberg2015STN} for image-style data transformation. $I_{last}$ refers to the past image-style data in Memory Bank and $I_f$ is the fused data of $I_{last}$ and current $I_{cur}$.
$Fusion(\cdot)$ can be {\it Add}, {\it Max}, {\it Concat} and {\it GRU-like}, as shown in Fig.~\ref{fig:SeqFusion}.

\subsection{SeqSampler}
\label{sec:SeqSampler}
SeqSampler is the key to perform the training of INT in an infinite-frames manner. It is designed to split original sequences to target length, and then generate the indices of them orderly. If the sequence is infinite-long, the training or inference can go on infinitely. DTSL is formed by executing SeqSampler with different target lengths for each epoch.

\begin{figure*}
    \centering
    \includegraphics[width=0.75\linewidth]{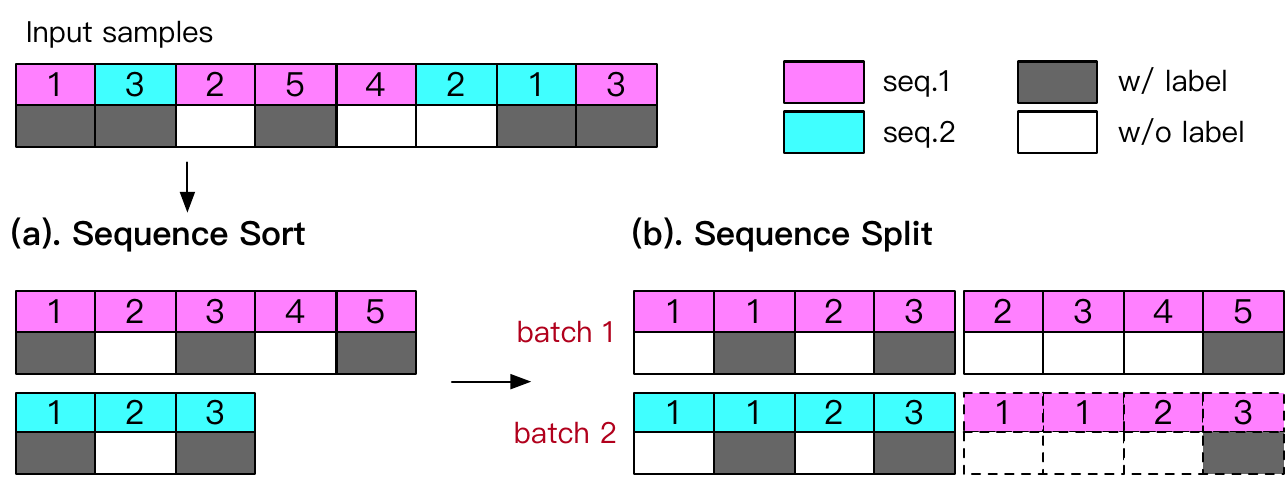}
    \caption{An example of SeqSampler. There are two sequences: seq1 contains 5 frames and seq2 has 3, both are interval labeled. Given the desired batch size 2 and target length 4, we need to get the final iteration indices. First, the two sequences are sorted separately. Then, the original sequences are splitted to 3 segments in the target length, and a segment is randomly replicated (dashed rectangles) to guarantee that both batches have the same number of samples.} 
    \label{fig:SeqSampler}
\end{figure*}

The length of original sequences in a dataset generally varies, for as in the Waymo Open Dataset~\cite{sun2020scalability}, where sequence lengths oscillate around 200. Certain datasets, such as nuScenes~\cite{caesar2020nuscenes}, may be interval labeled, with one frame labeled every ten frames. As a result, the SeqSampler should be designed with the idea that the source sequence will be non-fixed in length and will be annotated at intervals. The procedures of SeqSampler are as simple as \textit{Sequence Sort} and \textit{Sequence Split}, as indicated in Fig.~\ref{fig:SeqSampler}. In \textit{Sequence Sort}, we rearrange the random input samples orderly by sequences. Then split them to target length in \textit{Sequence Sort}, and may padding some of them to meet the batch or iteration demands.

\subsection{SeqAug}
\label{sec:SeqAug}
Data augmentation has been successful in many recent 3D detectors~\cite{2018SECOND,2019PointPillars,2021Center}. However, because of the shift in training paradigm, our suggested INT framework can not directly migrate current validated data augmentation methods. One of the main reasons for this is that INT is trained on a stream with a clear association between the before and after frames, whereas data augmentation is typically random, and the before and after frames could take various augmentation procedures. To solve this problem and allow INT to benefit from data augmentation, we must verify that a certain method of data augmentation on the same stream maintains the same random state at all times. We term the data augmentation that meets this condition SeqAug.
According to the data augmentation methods widely employed in pointcloud detection, SeqAug can be split into two categories: Sequence Point Transformation (flipping, rotation, translation, scaling, and so on) and Sequence GtAug (copy and paste of the ground truth pointclouds).

{\bf Sequence Point Transformation.} If a pointcloud is successively augmented by flipping $T_f$, rotation $T_r$, scaling $T_s$, and translation $T_t$, the other frames in the same stream must keep the same random state to establish a reasonable temporal relationship. In addition, because of these transformations, $T_{rel}$ in Eq.~\ref{eq:Tr} must be recalculated:
\begin{equation}\label{eq:Tr_aug}
    T_{rel} = T_t\cdot T_s \cdot T_r \cdot T_f \cdot T_{cur}^{-1} \cdot T_{last}  \cdot T_f \cdot T_r^{-1} \cdot T_s^{-1} \cdot T_t^{-1}
\end{equation}
where $T_{last}$ and $T_{cur}$ are the last and current frame's ego vehicle poses.

\begin{figure*}
    \centering
    \includegraphics[width=1.0\linewidth]{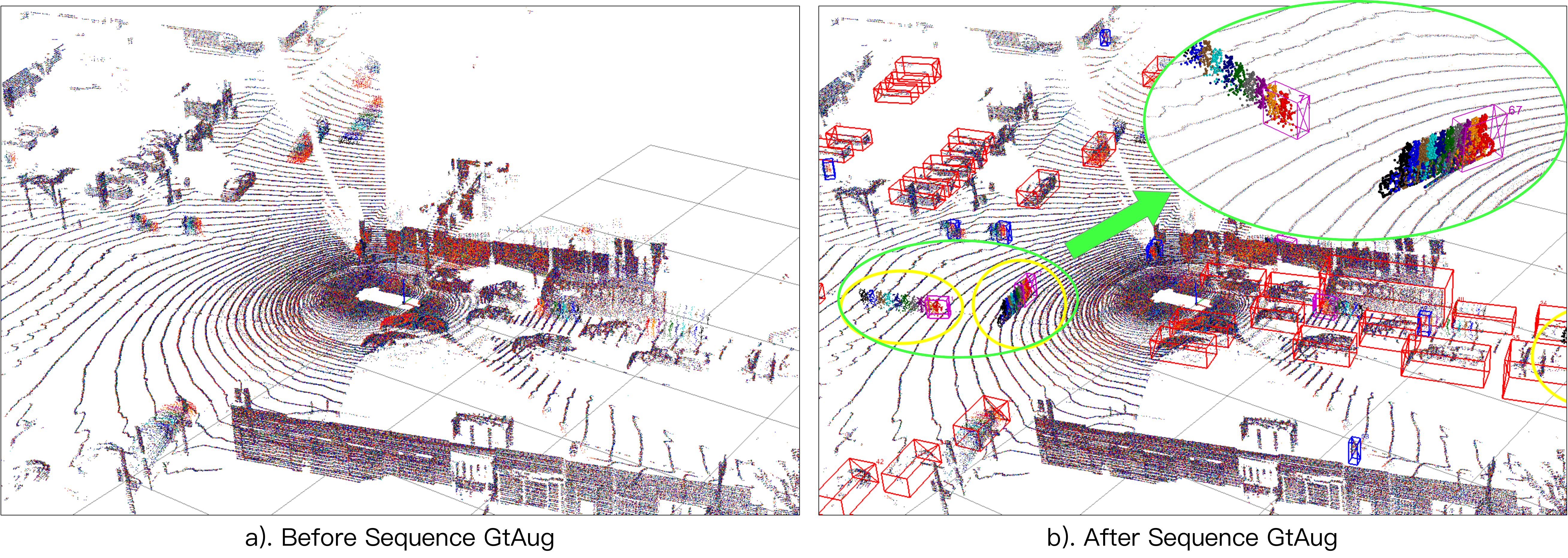}
    \caption{An example of Sequence GtAug. The colors of pointclouds represent different moments, with red being the current frame. }
    \label{fig:SeqAug}
\end{figure*}

{\bf Sequence GtAug.} Similarly, recording random states of the same stream is required to ensure that the same samples are fetched in the Gt database and that a specific Gt object can be copied and pasted consecutively, as shown in Fig.~\ref{fig:SeqAug}.

\section{Experiments}
In this paper, we built the proposed INT framework based on the highly competitive and popular detector CenterPoint~\cite{2021Center}.
In the following sections, we first briefly introduce the datasets in Sec.~\ref{sec:Datasets}, followed by a description of a few critical experimental setups in Sec.~\ref{sec:Experimental Settings}. The efficiency and effectiveness of the INT framework are then illustrated in Sec.~\ref{sec:Effe} by comparing it to the baseline CenterPoint, followed by Sec.~\ref{sec:compare_SOTA}, which compares the results of INT on the Waymo test set to other SOTAs. Finally, Sec.~\ref{sec:Ablation Studies} is several INT ablation experiments.
\subsection{Datasets}
\label{sec:Datasets}
This section briefly describes the two open datasets used in this paper.



{\bf Waymo Open Dataset.} Waymo~\cite{sun2020scalability} comprises 798, 202 and 150 sequences for train, validation and test, respectively. Each sequence lasts around 20 seconds and contains about 200 frames. There are three categories for detection: VEHICLE, PEDESTRIAN, and CYCLIST. The mean Average Precision (mAP) and mAP weighted by heading accuracy (mAPH) are the official 3D detection metrics. There are two degrees of difficulty: LEVEL 1 for boxes with more than five LiDAR points, and LEVEL 2 for boxes with at least one LiDAR point. In this paper, we utilize the officially prescribed mAPH on LEVEL 2 by default.

{\bf nuScenes.} There are 1000 driving sequences in nuScenes~\cite{caesar2020nuscenes}, with 700, 150, and 150 for training, validation, and testing, respectively. Each sequence lasts about 20 seconds and has a LiDAR frequency of 20 frames per second. The primary metrics for 3D detection are mean Average Precision (mAP) and nuScenes Detection Score (NDS). NDS is a weighted average of mAP and other attribute measurements such as translation, scale, orientation, velocity, and other box properties. In this study, we employ mAP and NDS as experimental results.

\subsection{Experimental Settings}
\label{sec:Experimental Settings}
We employ the same network designs and training schedules as CenterPoint~\cite{2021Center} and keep the positive and negative sample strategies, post-processing settings, loss functions, etc., unchanged.

{\bf Backbone settings.}
VoxelNet~\cite{2018SECOND,2018VoxelNet} and PointPillars~\cite{2019PointPillars} are two 3D encoders used by CenterPoint, dubbed CenterPoint-Voxel and CenterPoint-Pillar, respectively. Our INT also experiments with these two backbones, which correspond to INT-Voxel and INT-Pillar, respectively.

{\bf Frame settings.}
Although INT can be trained and inferred on an infinite number of frames, the sequence length of the actual dataset is finite. To facilitate comparison with previous work and demonstrate the benefits of the INT framework, we select a few specific frames on nuScenes and Waymo Open Dataset. We use 10, 20 and 100 training frames on nuScenes, and 10 training frames on Waymo Open Dataset.

{\bf Fusion settings.}
On INT, we choose three kinds of data added to Memory Bank: point-style foreground pointcloud, image-style intermediate feature map, and image-style final prediction map. For the foreground pointcloud, we fuse the historical points with the current points during the input phase and then update them based on the predictions at the end of network. For the intermediate feature map, we fuse and update its historical information at the same position before the Region Proposal Network. For the final prediction map, we fuse and update historical information simultaneously before the detection header. Appendix~\ref{app:INT-Voxel-framework} takes INT-Voxel as a typical example to provide a more specific explanation.

{\bf Latency settings.}
To test the network's actual latency, we remove redundant parts of the data processing in CenterPoint~\cite{2021Center} and just maintain the data IO and memory transfer (to the GPU) operations. We shift the essential voxelization component to the GPU to limit the CPU's influence. We also build up data prefetching in the dataloader to lessen IO effect and run latency tests when it is stabilized. Finally, the following test circumstances are used: CUDA Version 10.2, cudnn 7.6.5, GeForce RTX 2070 SUPER, Driver Version 460.91.03.

\subsection{Effectiveness and Efficiency}
\label{sec:Effe}
As shown in Table~\ref{table:effe_waymo} and~\ref{table:effe_nusc}, we first compare to the baseline CenterPoint to demonstrate the paper's main point, i.e., the effectiveness and efficiency of INT. In these two tables, we try two types of backbone, termed E-PointPillars and E-VoxelNet in the columns, and the unit of latency is milliseconds. The settings of INT can be referred to Table~\ref{table:AE_fusionObjects_waymo} and~\ref{table:AE_fusionObjects_nusc}. CenterPoint with multiple frames refers to concatenating multi-frame pointclouds at the input level, which introduces repetitive computation and additional memory burden as analyzed in Sec.~\ref{subsec:related_work_multi}. For example, two-frames CenterPoint in Table~\ref{table:effe_waymo} increases the latency by around 20 ms when compared to its single-frame counterpart (more details in Appendix~\ref{app:latency_details}). In contrast, the latency of INT is unaffected by the number of frames used, and its performance is much better than that of multi-frame CenterPoint as the number of frames grows. As can be obviously observed in Table~\ref{table:effe_waymo} and~\ref{table:effe_nusc}, INT shows significant improvements in both latency and performance, gaining around 7\% mAPH (Waymo) and 15\% NDS (nuScenes) boost while only adding 2\~{}4ms delay when compared to single-frame CenterPoint. 

\setlength{\tabcolsep}{3pt}
\begin{table}
\begin{center}
\caption{Effectiveness and efficiency of INT on Waymo Open Dataset {\it val} set. The APH of L2 difficulty is reported. The "-2s" suffix in the rows means two-stage model. CenterPoint's mAPH results are obtained from \href{https://github.com/tianweiy/CenterPoint/tree/master/configs/waymo}{official website}, except for those with a *, which are missing from the official results and were reproduced by us.}
\label{table:effe_waymo}
\resizebox{1.0\linewidth}{!}{%
\begin{tabular}{c|c|ccccc|ccccc}
\hline
\multirow{2}{*}{methods} & \multirow{2}{*}{frames} & \multicolumn{5}{c|}{E-PointPillars} & \multicolumn{5}{c}{E-VoxelNet} \\
 &  & VEH↑ & PED↑ & CYC↑ & mAPH↑ & latency↓ & VEH↑ & PED↑ & CYC↑ & mAPH↑ & latency↓ \\ \hline
CenterPoint & 1 & 65.5 & 55.1 & 60.2 & 60.3 & \textbf{57.7} & 66.2 & 62.6 & 67.6 & 65.5 & \textbf{71.7} \\
CenterPoint & 2 & 66.6* & 61.9* & 62.3* & 63.6* & 77.8 & 67.3 & 67.5 & 69.9 & 68.2 & 90.9 \\
INT (ours) & 2 & 66.2 & 60.4 & 64.4 & 63.7 & 61.6 & 69.4 & 69.1 & 72.6 & 70.3 & 74.0 \\
INT (ours) & 10 & \textbf{69.6} & \textbf{66.3} & \textbf{65.7} & \textbf{67.2} & 61.6 & \textbf{72.2} & \textbf{72.1} & \textbf{75.3} & \textbf{73.2} & 74.0 \\ \hline
CenterPoint-2s & 1 & 66.7 & 55.9 & 61.7 & 61.4 & \textbf{61.7} & 67.9 & 65.6 & 68.6 & 67.4 & \textbf{76.6} \\
CenterPoint-2s & 2 & 68.4* & 63.0* & 64.3* & 65.2* & 82.9 & 69.7 & 70.3 & 70.9 & 70.3 & 95.8 \\
INT-2s (ours) & 2 & 67.9 & 61.7 & 66.0 & 65.2 & 65.9 & 70.8 & 68.7 & 73.1 & 70.8 & 78.9 \\
INT-2s (ours) & 10 & \textbf{70.8} & \textbf{67.0} & \textbf{68.1} & \textbf{68.6} & 65.9 & \textbf{73.3} & \textbf{71.9} & \textbf{75.6} & \textbf{73.6} & 78.9 \\ \hline
\end{tabular}
}%
\end{center}
\end{table}
\setlength{\tabcolsep}{1.4pt}

\setlength{\tabcolsep}{4pt}
\begin{table}
\begin{center}
\caption{Effectiveness and efficiency of INT on nuScenes {\it val} set. CenterPoint's mAP and NDS results are obtained from \href{https://github.com/tianweiy/CenterPoint/tree/master/configs/nusc}{official website}, except for those with a *, which are missing from the official results and were reproduced by us.}
\label{table:effe_nusc}
\resizebox{0.8\linewidth}{!}{%
\begin{tabular}{c|c|ccc|ccc}
\hline
\multirow{2}{*}{methods} & \multirow{2}{*}{frames} & \multicolumn{3}{c|}{E-PointPillars} & \multicolumn{3}{c}{E-VoxelNet} \\
 &  & mAP↑ & NDS↑ & latency↓ & mAP↑ & NDS↑ & latency↓ \\ \hline
CenterPoint & 1 & 42.5* & 46.4* & \textbf{39.2} & 49.7* & 50.7* & \textbf{81.1} \\
CenterPoint & 10 & 50.3 & 60.2 & 49.4 & 59.6 & 66.8 & 117.2 \\
INT (ours) & 10 & 49.3 & 59.9 & 43.0 & 58.5 & 65.5 & 84.1 \\
INT (ours) & 20 & 50.7 & 61.0 & 43.0 & 60.9 & 66.9 & 84.1 \\
INT (ours) & 100 & \textbf{52.3} & \textbf{61.8} & 43.0 & \textbf{61.8} & \textbf{67.3} & 84.1 \\ \hline
\end{tabular}
}%
\end{center}
\end{table}
\setlength{\tabcolsep}{1.4pt}


\subsection{Comparison with SOTAs}
\label{sec:compare_SOTA}
In this section, we compare the results of INT on the Waymo {\it test} set with those of other SOTAs schemes, as shown in Table~\ref{table:sota_waymo}. The approaches are divided into two groups in the table: single-frame and multi-frame. It is seen that the multi-frame scheme is generally superior to the single-frame scheme. Most multi-frame approaches employ the original pointclouds concatenation~\cite{2021RSN,2021Center,2021AFDetV2}, and we can see that the number of frames is used fewer due to computational and memory constraints. Finally, our suggested INT scheme outperforms other SOTA schemes by a large margin. As far as we know, INT is the best non-ensemble approach on Waymo Open Dataset leaderboard\footnote{\url{https://waymo.com/open/challenges/2020/3d-detection/}\label{leadborad}}.

\setlength{\tabcolsep}{4pt}
\begin{table}
\begin{center}
\caption{Comparison with SOTAs on Waymo Open Dataset {\it test} set. We only present the non-emsemble approaches, and INT is currently the best non-emsemble solution on the Waymo Open Dataset leaderboard\textsuperscript{\ref{leadborad}}, to the best of our knowledge. Accessed on 2 March 2022.}
\label{table:sota_waymo}
\resizebox{0.9\linewidth}{!}{%
\begin{tabular}{c|c|cc|cc|cc|cc}
\hline
\multirow{2}{*}{Methods} & \multirow{2}{*}{Frames} & \multicolumn{2}{c|}{VEH-APH↑} & \multicolumn{2}{c|}{PED-APH↑} & \multicolumn{2}{c|}{CYC-APH↑} & \multicolumn{2}{c}{mAPH↑} \\
 &  & L1 & L2 & L1 & L2 & L1 & L2 & L1 & L2 \\ \hline
StarNet~\cite{2019StarNet} & 1 & 61.0 & 54.5 & 59.9 & 54.0 & - & - & - & - \\
PointPillars~\cite{2019PointPillars} & 1 & 68.1 & 60.1 & 55.5 & 50.1 & - & - & - & - \\
RCD~\cite{2020RCD} & 1 & 71.6 & 64.7 & - & - & - & - & - & - \\
M3DeTR~\cite{2021M3DeTR} & 1 & 77.2 & 70.1 & 58.9 & 52.4 & 65.7 & 63.8 & 67.1 & 61.9 \\
HIK-LiDAR~\cite{2021CenterAtt} & 1 & 78.1 & 70.6 & 69.9 & 64.1 & 69.7 & 67.2 & 72.6 & 67.3 \\
CenterPoint~\cite{2021Center} & 1 & 79.7 & 71.8 & 72.1 & 66.4 & - & - & - & - \\ \hline
3D-MAN~\cite{yang20213D-MAN} & 15 & 78.3 & 70.0 & 66.0 & 60.3 & - & - & - & - \\
RSN~\cite{2021RSN} & 3 & 80.3 & 71.6 & 75.6 & 67.8 & - & - & - & - \\
CenterPoint~\cite{2021Center} & 2 & 80.6 & 73.0 & 77.3 & 71.5 & 73.7 & 71.3 & 77.2 & 71.9 \\
CenterPoint++~\cite{2021Center} & 3 & 82.3 & 75.1 & 78.2 & 72.4 & 73.3 & 71.1 & 78.0 & 72.8 \\
AFDetV2~\cite{2021AFDetV2} & 2 & 81.2 & 73.9 & 78.1 & 72.4 & 75.4 & 73.0 & 78.2 & 73.1 \\
INT (ours) & 10 & 83.1 & 76.2 & 78.5 & 72.8 & 74.8 & 72.7 & 78.8 & 73.9 \\
INT (ours) & 100 & \textbf{84.3} & \textbf{77.6} & \textbf{79.7} & \textbf{74.0} & \textbf{76.3} & \textbf{74.1} & \textbf{80.1} & \textbf{75.2} \\ \hline
\end{tabular}
}%
\end{center}
\end{table}
\setlength{\tabcolsep}{1.4pt}

\subsection{Ablation Studies}
\label{sec:Ablation Studies}

\subsubsection{Impact of different fusion data.}
\label{sec:AE_fusionObjects}
This section investigates the impact of various fusion data used in INT. We use one kind of point-style data, the foreground pointcloud, and two kinds of image-style data, the intermediate feature map before RPN and the final prediction map in this paper. The fusion method of foreground pointcloud is termed as {\it PC Fusion} which is explained in Sec.~\ref{subsec:point_style}. The fusion method of  the intermediate feature map and the final prediction map is {\it Concat}, as described in Sec.~\ref{subsec:image_style}, named as {\it FM Fusion} and {\it PM Fusion}, respectively. The fusion results of these three kinds of data on Waymo Open Dataset and nuScenes are shown in Table \ref{table:AE_fusionObjects_waymo} and \ref{table:AE_fusionObjects_nusc}. First, the table's performance column shows that all the three fusion data have considerable performance boosts, with {\it PC Fusion} having the highest effect gain. Then according to the latency column, the increase in time of different fusion data is relatively small, which is very cost-effective given the performance benefit.

\setlength{\tabcolsep}{4pt}
\begin{table}
\begin{center}
\caption{Impact of different fusion data on Waymo Open Dataset{\it val} set. By default, the training sequence length was set to 10 frames. In order to classify how the final result comes in Table~\ref{table:effe_waymo}, we also add a column called "Two Stage".}
\label{table:AE_fusionObjects_waymo}
\resizebox{1.0\linewidth}{!}{%
\begin{tabular}{cccc|ccccc|ccccc}
\hline
\multirow{2}{*}{\begin{tabular}[c]{@{}c@{}}PC\\ Fusion\end{tabular}} & \multirow{2}{*}{\begin{tabular}[c]{@{}c@{}}FM\\ Fusion\end{tabular}} & \multirow{2}{*}{\begin{tabular}[c]{@{}c@{}}PM\\ Fusion\end{tabular}} & \multirow{2}{*}{\begin{tabular}[c]{@{}c@{}}Two\\ Stage\end{tabular}} & \multicolumn{5}{c|}{E-PointPillars} & \multicolumn{5}{c}{E-VoxelNet} \\
 &  &  &  & VEH↑ & PED↑ & CYC↑ & mAPH↑ & latency↓ & VEH↑ & PED↑ & CYC↑ & mAPH↑ & latency↓ \\ \hline
 &  &  &  & 65.5 & 55.1 & 60.2 & 60.3 & \textbf{57.4} & 66.2 & 62.6 & 67.6 & 65.5 & \textbf{71.5} \\
$\surd$ &  &  &  & 68.1 & 65.8 & 65.4 & 66.4 & 59.5 & 71.7 & 70.8 & 74.2 & 72.3 & 72.7 \\
 & $\surd$ &  &  & 63.6 & 63.3 & 64.7 & 63.8 & 59.0 & 66.1 & 67.3 & 73.8 & 69.1 & 72.2 \\
 &  & $\surd$ &  & 66.4 & 64.0 & 64.2 & 64.5 & 58.2 & 67.7 & 68.1 & 74.1 & 70.0 & 72.0 \\
$\surd$ & $\surd$ &  &  & 69.5 & 66.8 & 64.8 & 67.0 & 60.9 & 72.0 & 71.8 & 76.5 & 73.5 & 73.3 \\
$\surd$ & $\surd$ & $\surd$ &  & 69.6 & 66.3 & 65.7 & 67.2 & 61.6 & 72.2 & \textbf{72.1} & \textbf{76.1} & 73.5 & 74.0 \\
$\surd$ & $\surd$ & $\surd$ & $\surd$ & \textbf{70.8} & \textbf{67.0} & \textbf{68.1} & \textbf{68.6} & 65.9 & \textbf{73.3} & 71.9 & 75.6 & \textbf{73.6} & 78.9 \\ \hline
\end{tabular}
}%
\end{center}
\end{table}
\setlength{\tabcolsep}{1.4pt}


\setlength{\tabcolsep}{4pt}
\begin{table}
\begin{center}
\caption{Impact of different fusion data on nuScenes {\it val} set. By default, the training sequence length was set to 10 frames. In order to classify how the final result comes in Table~\ref{table:effe_nusc}, we also add a column called "100 frames".}
\label{table:AE_fusionObjects_nusc}
\resizebox{0.7\linewidth}{!}{%
\begin{tabular}{cccc|ccc|ccc}
\hline
\multirow{2}{*}{\begin{tabular}[c]{@{}c@{}}PC\\ Fusion\end{tabular}} & \multirow{2}{*}{\begin{tabular}[c]{@{}c@{}}FM\\ Fusion\end{tabular}} & \multirow{2}{*}{\begin{tabular}[c]{@{}c@{}}PM\\ Fusion\end{tabular}} & \multirow{2}{*}{\begin{tabular}[c]{@{}c@{}}100\\ frames\end{tabular}} & \multicolumn{3}{c|}{E-PointPillars} & \multicolumn{3}{c}{E-VoxelNet} \\
 &  &  &  & mAP & NDS & latency & mAP & NDS & latency \\ \hline
 &  &  &  & 42.5 & 46.4 & \textbf{39.2} & 49.7 & 50.7 & \textbf{81.1} \\
$\surd$ &  &  &  & 47.1 & 58.5 & 41.2 & 56.6 & 64.5 & 82.4 \\
 & $\surd$ &  &  & 48.4 & 57.4 & 40.5 & 55.9 & 56.6 & 82.1 \\
 &  & $\surd$ &  & 45.3 & 56.0 & 39.7 & 53.9 & 62.7 & 82.0 \\
$\surd$ & $\surd$ &  &  & 48.7 & 59.8 & 42.4 & 58.4 & 65.2 & 83.3 \\
$\surd$ & $\surd$ & $\surd$ &  & 49.3 & 59.9 & 43.0 & 58.5 & 65.5 & 84.1 \\
$\surd$ & $\surd$ & $\surd$ & $\surd$ & \textbf{52.3} & \textbf{61.8} & 43.0 & \textbf{61.8} & \textbf{67.3} & 84.1 \\ \hline
\end{tabular}
}%
\end{center}
\end{table}
\setlength{\tabcolsep}{1.4pt}


\subsubsection{Impact of training sequence length.}
\label{sec:AE_frames_waymo}
 The length indicated here actually refer to the maximum length since Dynamic Training Sequence Length (DTSL) is used in training. Fig.~\ref{fig:frames-exp-waymo} depicts the relationship between frames and performance for the 2-stage INT-Voxel model on the Waymo Open Dataset. We also plot the 2-stage CenterPoint-Voxel results together to make comparisons clearer. As shown in Fig.~\ref{fig:frames-exp-waymo}(b), INT improves as the number of frames increases, although there is saturation after a certain point (See Appendix~\ref{app:saturation_explanation} for more explanation); as for Fig.~\ref{fig:frames-exp-waymo}(c), the time consumed by INT is slightly higher than that of single-frame CenterPoint, but it does not increase with the number of frames.

\subsubsection{Impact of sequence augmentation.}
\label{sec:AE_aug}
Sec.~\ref{sec:SeqAug} introduces SeqAug, a data augmentation technique for on-stream training, and this section examines the role of Point Transformation and GtAug in SeqAug. As seen in Table~\ref{table:AE_aug}, both augmentation strategies result in significant performance improvements, making data augmentation essential for INT training just as regular detectors do.

\setlength{\tabcolsep}{3pt}
\begin{table}
\begin{center}
\caption{Impact of SeqAug on Waymo Open Dataset {\it val} set. One-stage INT-Pillar and INT-Voxel are used.}
\label{table:AE_aug}
\resizebox{1.0\linewidth}{!}{%
\begin{tabular}{cc|cccc|cccc}
\hline
\multirow{2}{*}{\begin{tabular}[c]{@{}c@{}}Sequence\\ Point Trans.\end{tabular}} & \multirow{2}{*}{\begin{tabular}[c]{@{}c@{}}Sequence\\ GtAug\end{tabular}} & \multicolumn{4}{c|}{E-PointPillars} & \multicolumn{4}{c}{E-VoxelNet} \\
 &  & VEL-APH↑ & PED-APH↑ & CYC-APH↑ & mAPH↑ & VEL-APH↑ & PED-APH↑ & CYC-APH↑ & mAPH↑ \\ \hline
 &  & 59.6 & 56.7 & 56.2 & 57.5 & 65.0 & 62.4 & 61.9 & 63.1 \\
$\surd$ &  & 69.3 & 65.0 & 62.4 & 65.6 & 71.8 & 71.4 & 73.1 & 72.1 \\
$\surd$ & $\surd$ & \textbf{69.6} & \textbf{66.3} & \textbf{65.7} & \textbf{67.2} & \textbf{72.2} & \textbf{72.1} & \textbf{76.1} & \textbf{73.5} \\ \hline
\end{tabular}
}%
\end{center}
\end{table}
\setlength{\tabcolsep}{1.4pt}



\subsubsection{More ablation studies in appendix.}
\label{sec:AE_more}
The impact of sequence length on nuScenes is shown in Appendix~\ref{app:AE_frames_nusc}. The performance and latency of temporal fusion methods for image-style data (proposed in Sec.~\ref{subsec:image_style}) are shown in Appendix~\ref{app:AE_fusionMethods}. The impact of DTSL proposed in Sec.~\ref{subsec:DTSL} can be found in Appendix~\ref{app:AE_dtsl}.

\section{Conclusion}
In this paper, we present INT, a novel on-stream training and prediction framework that, in theory, can employ an infinite number of frames while using about the same amount of computational and memory cost as a single-frame detector. To make INT feasible, we propose three key modules, i.e., SeqFusion, SeqSampler, and SeqAug. We utilize INT on the popular CenterPoint, with significant latency reductions and performance improvements, and rank  {\bf 1st} currently on Waymo Open Dataset 3D Detection leaderboard among the non-ensemble SOTA methods. Moreover, the INT is a general multi-frame system, which may be used for tasks like segmentation and motion as well as detection.


\setlength{\parskip}{0.5em}\noindent{\bf Acknowledgement:}
This work was supported by Alibaba Group through Alibaba Innovative Research (AIR)
Program and Alibaba Research Intern Program.



\par\vfill\par

\clearpage
%
%
\bibliographystyle{splncs04}
\bibliography{egbib}

\clearpage

\appendix

\section{Appendix}

\subsection{Impact of training sequence length on nuScenes.}
\label{app:AE_frames_nusc}
\begin{figure*}
    \centering
    \includegraphics[width=0.98\linewidth]{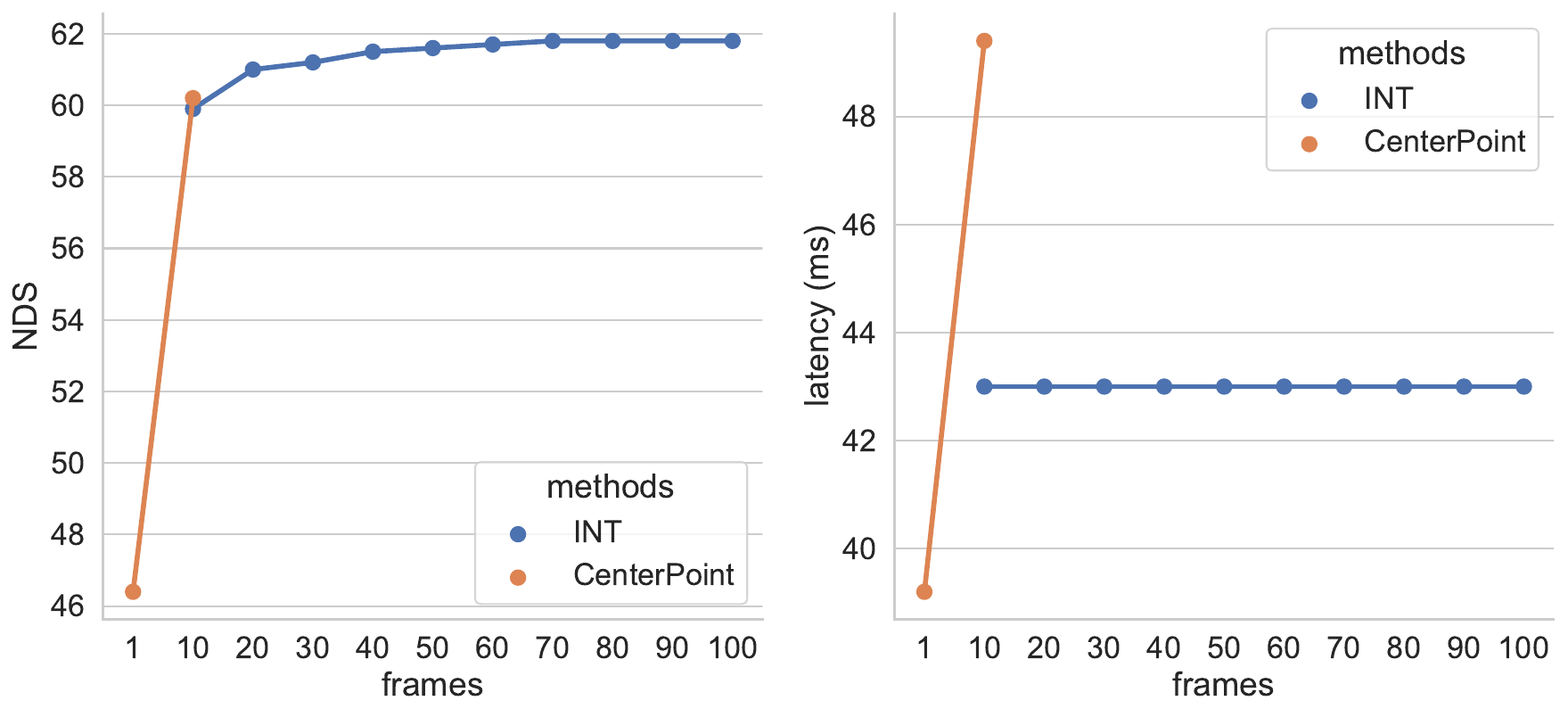}
    \caption{Impact of training frame length on nuScenes {\it val} set. While CenterPoint's performance improves as the number of frames grows, the latency also increases dramatically. On the other hand, our INT keeps the same latency while increasing performance.}
    \label{fig:frames-exp-nusc}
\end{figure*}

In Sec.~{\ref{sec:AE_frames_waymo}} {\bf Impact of training sequence length} of main paper, we have showed the results on Waymo Open Dataset~\cite{sun2020scalability}. This section we show the results on nuScenes~\cite{caesar2020nuscenes}, as depicted in Figure~\ref{fig:frames-exp-nusc}, which depicts the relationship between frames and performance for the proposed INT-Pillar and CenterPoint-Pillar~\cite{2021Center} model. Looking at the left panel first, INT performance improves as the number of frames increases, although there is saturation after a certain point; as for the right panel, the time consumed by INT is slightly higher than that of single-frame CenterPoint, but it does not increase with the number of frames.

\subsection{Impact of fusion methods for image-style data.}
\label{app:AE_fusionMethods}
As indicated in {\bf images-style fusion} of Sec.~\ref{sec:SeqFusion} in the main paper, we propose four fusion algorithms for image-style data. This section examines the performance and latency of these fusion methods. As shown in Table~\ref{table:AE_fusionMethods}, the {\it Concat} and {\it GRU-like} solutions provide better performance, whereas {\it Add} and {\it Max} require less amount of latency. Taking both performance and efficiency into account, we choose the {\it Concat} as the default temporal fusion method for image-style data fusion experiments in both the main paper and the appendix. To accelerate experiments in Table~\ref{table:AE_fusionMethods}, we use 1/4 training sequences for both Waymo Open Dataset and nuScenes, and we use 1/4 validation sequences for Waymo and the whole validation sequences for nuScenes.

\setlength{\tabcolsep}{4pt}
\begin{table}
\begin{center}
\caption{Impact of different fusion methods for image-style data on Waymo Open Dataset~\cite{sun2020scalability} and nuScenes~\cite{caesar2020nuscenes} {\it val} set. INT-Pillars is employed in these experiments. To accelerate experiments, only 1/4 training sequences are used.}
\label{table:AE_fusionMethods}
\resizebox{0.9\linewidth}{!}{%
\begin{tabular}{c|ccc|ccccc}
\hline
\multirow{2}{*}{\begin{tabular}[c]{@{}c@{}}Fusion\\ methods\end{tabular}} & \multicolumn{3}{c|}{nuScenes} & \multicolumn{5}{c}{Waymo} \\
 & mAP↑ & NDS↑ & latency↓ & VEL↑ & PED↑ & CYC↑ & mAPH↑ & latency↓ \\ \hline
Baseline & 26.8 & 35.5 & \textbf{39.2} & 59.1 & 48.6 & 41.1 & 49.6 & \textbf{57.4} \\
Add & 31.5 & 42.0 & 39.6 & 58.9 & 54.6 & 54.8 & 56.1 & 58.3 \\
Concat & \textbf{33.7} & \textbf{45.3} & 40.5 & \textbf{59.8} & 58.3 & 62.5 & \textbf{60.2} & 59.0 \\
Max & 30.3 & 42.9 & 39.6 & 58.7 & 54.9 & 55.1 & 56.2 & 58.4 \\
GRU-like & 31.9 & 44.8 & 41.0 & 58.1 & \textbf{58.9} & \textbf{63.8} & \textbf{60.2} & 59.7 \\ \hline
\end{tabular}
}%
\end{center}
\end{table}
\setlength{\tabcolsep}{1.4pt}

\subsection{Impact of Dynamic Training Sequence Length (DTSL).}
\label{app:AE_dtsl}
{\bf Dynamic Training Sequence Length (DTSL)} in Sec 3.1 of main paper explains why DTSL is required for INT training, and this part confirms DTSL's function. Training a 100-frame INT-Pillars network on nuScenes improves the NDS with DTSL by 2.5\% compared to no DTSL, as demonstrated in Table~\ref{table:AE_dtsl}.

\setlength{\tabcolsep}{3pt}
\begin{table}
\begin{center}
\caption{Impact of Dynamic Training Sequence Length (DTSL) on nuScenes {\it val} set.}
\label{table:AE_dtsl}
\resizebox{0.3\linewidth}{!}{%
\begin{tabular}{c|cc}
\hline
 & mAP & NDS \\ \hline
w/o DTSL & 50.2 & 59.3 \\
w/ DTSL & \textbf{52.3} & \textbf{61.8} \\ \hline
\end{tabular}
}%
\end{center}
\end{table}
\setlength{\tabcolsep}{1.4pt}

\subsection{A typical example of INT-Voxel framework.}
\begin{figure*}
    \centering
    \includegraphics[width=1.0\linewidth]{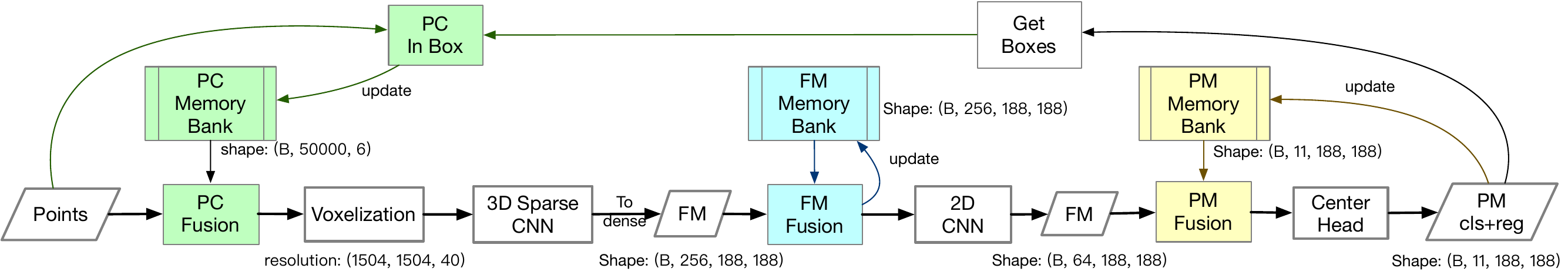}
    \caption{Pipeline of INT-VoxelNet on Waymo. PC, FM and PM represent Point Cloud, Feature Map and Prediction Map, respectively.}
    \label{fig:INT-Voxel-framework}
\end{figure*}

The point-style data in MB is initialized as a $B\times N\times C$ tensor, and image-style data is a $B\times C\times H\times W$ tensor, where $B$ is batch size, $C$ is feature dimension, $N$ is points number (50,000 in the paper),  $H/W$ is the feature map size. Here we take INT-Voxel as an example, and Fig.~\ref{fig:INT-Voxel-framework} shows the positions of fusion modules, which are commonly described in \textbf{Fusion settings} of Sec.~\ref{sec:Experimental Settings}.
\label{app:INT-Voxel-framework}

\subsection{Explanation on performance saturation.}
\label{app:saturation_explanation}
From Fig.~\ref{fig:frames-exp-waymo} and~\ref{fig:frames-exp-nusc}, we can see a performance saturation while frames increases. The reason for this phenomenon is that:
(1) Point-style data in MB has a limited number of points (current 50,000), so points that are too old do not remain in the MB any more. Enlarging the point MB or improving the update policy may ease the problem.
(2) Image-style data is exponentially decaying in current fusion methods, so the contribution of too old information is almost negligible. Improving the fusion methods may alleviate this problem.

\subsection{Latency details for CenterPoint in Table~\ref{table:effe_waymo}}
\label{app:latency_details}
As illustrated in \textbf{Latency settings} of Sec.~\ref{sec:Experimental Settings}, we put the voxelization to GPU. The voxelization implementation follows \href{https://github.com/open-mmlab/OpenPCDet/blob/master/pcdet/models/backbones_3d/vfe/dynamic_pillar_vfe.py}{OpenPCDet} for E-Pillar, and \href{https://github.com/tianweiy/CenterPoint/blob/new_release/det3d/models/readers/dynamic_voxel_encoder.py}{CeneterPoint} for E-Voxel. Table~\ref{table:latency_details_cp-pillar} and~\ref{table:latency_details_cp-voxel} show the detailed cost of each module in CenterPoint~\cite{2021Center}.
\setlength{\tabcolsep}{5pt}
\begin{table}
\begin{center}
\caption{Detailed latency of CenterPoint-Pillar on Waymo Open Dataset.}
\label{table:latency_details_cp-pillar}
\resizebox{0.8\linewidth}{!}{%
\begin{tabular}{c|ccccc}
\hline
latency(ms) & \begin{tabular}[c]{@{}l@{}}copy to\\ GPU\end{tabular} & voxelization & \begin{tabular}[c]{@{}l@{}}MLP +\\ scatter\_to\_bev\end{tabular} & \begin{tabular}[c]{@{}l@{}}2D CNN\\ + post\end{tabular} & total \\ \hline
E-Pillar 1f & 0.6 & 1.8 & 5.5 & 49.8 & 57.7 \\
E-Pillar 2f & 1.4 (+0.8) & 11.4 (+9.6) & 14.6 (+9.1) & 50.4 (+0.6) & 77.8 (+20.1) \\ \hline
\end{tabular}
}%
\end{center}
\end{table}
\setlength{\tabcolsep}{5pt}
\begin{table}[h]
\begin{center}
\caption{Detailed latency of CenterPoint-Voxel on Waymo Open Dataset.}
\label{table:latency_details_cp-voxel}
\resizebox{0.8\linewidth}{!}{%
\begin{tabular}{c|ccccc}
\hline
latency(ms) & \begin{tabular}[c]{@{}c@{}}copy to\\ GPU\end{tabular} & voxelization & sparse conv & \begin{tabular}[c]{@{}c@{}}2D CNN \\ + post\end{tabular} & total \\ \hline
E-Voxel 1f & 0.6 & 1.9 & 43.6 & 25.6 & 71.7 \\
E-Voxel 2f & 1.3 (+0.7) & 3.7 (+1.8) & 60.1 (+16.5) & 25.8 (+0.2) & 90.9 (+19.2) \\ \hline
\end{tabular}
}%
\end{center}
\end{table}

\subsection{Prediction Visualizations}
\begin{figure*}
    \centering
    \includegraphics[width=1.0\linewidth]{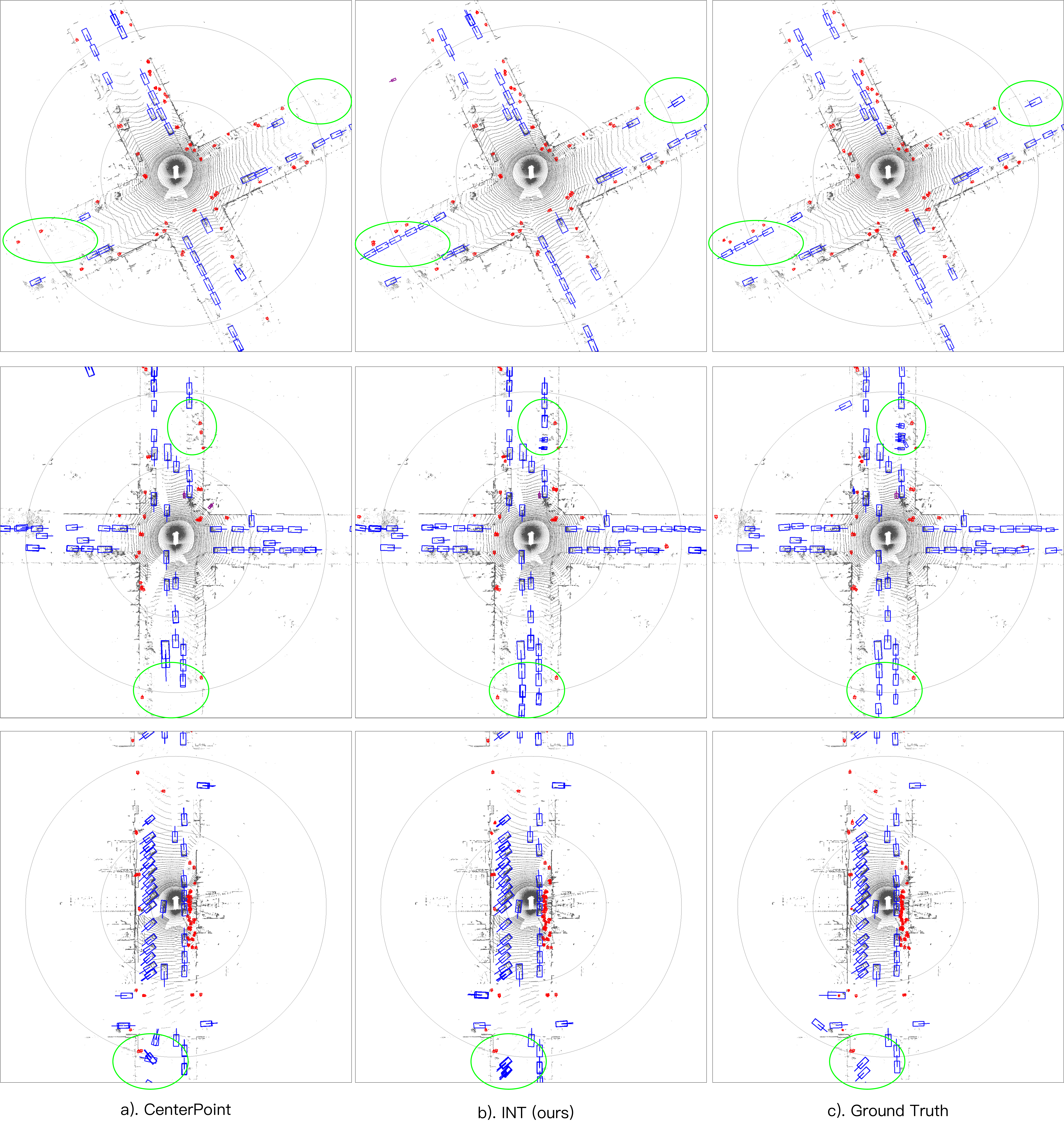}
    \caption{The detection results on Waymo Open Dataset {\it val} set. The bounding boxes of VEHICLE, PEDESTRIAN and CYCLIST are in the color blue, red and magenta respectively. As pointed out in green circles, INT is obviously more advantageous in detecting long-range targets because it has more historical information.}
    \label{fig:viz}
\end{figure*}

We compared the prediction results of 2-stage CenterPoint-Voxel and INT-Voxel on Waymo Open Dataset {\it val} set in Figure~\ref{fig:viz}.

\end{document}